\documentclass{article}
\usepackage[preprint]{colm2026_conference}

\usepackage{microtype}
\usepackage{hyperref}
\usepackage{url}
\usepackage{graphicx}
\usepackage{booktabs}
\usepackage{amsmath}
\usepackage{amssymb}
\usepackage{algorithm}
\usepackage{algorithmic}
\usepackage{multirow}
\usepackage{xspace}
\usepackage{tikz}
\usetikzlibrary{calc}
\usepackage{lineno}

\usepackage{amsmath,amsfonts,bm}

\def\eqref#1{equation~\ref{#1}}

\def\1{\bm{1}}

\DeclareMathAlphabet{\mathsfit}{\encodingdefault}{\sfdefault}{m}{sl}
\SetMathAlphabet{\mathsfit}{bold}{\encodingdefault}{\sfdefault}{bx}{n}

\definecolor{darkblue}{rgb}{0, 0, 0.5}
\hypersetup{colorlinks=true, citecolor=darkblue, linkcolor=darkblue, urlcolor=darkblue}

\newcommand{\robophd}{RoboPhD\xspace}
\newcommand{\koth}{KotH\xspace}

\title{RoboPhD: Evolving Diverse Complex Agents\\Under Tight Evaluation Budgets}

\author{Andrew Borthwick \& Stephen Ash \& Anthony Galczak \\
Independent Researchers\\
\texttt{\{AEBorthwick, stevemash, wgalczak\}@gmail.com}
}

\begin{document}

\maketitle

\begin{abstract}
2026 has brought an explosion of interest in LLM-guided evolution of agentic artifacts, with systems like GEPA and Autoresearch demonstrating that LLMs can iteratively improve prompts, code, and agent architectures across diverse domains. As adoption accelerates, a central question emerges: given the same information, the same seed agent, and the same objective, which optimization algorithm yields the best results under the same evaluation budget?  This question becomes critical when evaluations are expensive, such as when they require human judgment or multiple LLM calls.

We present the first systematic comparison of three optimization paradigms---Elo tournament selection (\robophd), Pareto-based selection (GEPA), and greedy hill-climbing (Autoresearch)---across four benchmarks spanning abstract reasoning, cloud scheduling, database query generation, and financial document QA, all under a fixed budget of 1{,}500 evaluations. \robophd introduces validation-free evolution: instead of splitting the budget between training and validation, it uses Elo competition on training data to simultaneously evaluate agents and drive evolution. All three systems receive seed agents with diagnostic \texttt{print()} statements that evolution can grow, enabling \emph{self-instrumenting agents} that develop increasingly informative diagnostics for the benefit of their evolutionary successors.

Using a single default configuration, \robophd outperforms both GEPA and Autoresearch on three of four benchmarks, losing only on the simplest task, where the winning solution (from our Autoresearch adaptation) required under 90 lines of code. On ARC-AGI, \robophd evolves a 22-line seed agent into a 1{,}013-line multi-strategy system, improving accuracy from 27.8\% to 65.8\% using Gemini 3.1 Flash Lite as the solver. We release \robophd as a versatile toolkit under the MIT license with a simple \texttt{optimize\_anything()} API for evolving diverse complex agents.
\end{abstract}

\section{Introduction}
\label{sec:intro}

The automatic evolution of AI agents and code artifacts has emerged as a rapidly growing research direction. GEPA's \texttt{optimize\_anything()} \citep{gepa2025,gepa_oa2026} and Karpathy's Autoresearch \citep{autoresearch2026} demonstrate that large language models can iteratively
improve text artifacts---from prompts and agent architectures
\citep{gepa2025,gepa_oa2026} to training code and model
configurations \citep{autoresearch2026}---by analyzing
evaluation feedback and proposing refinements.
This direction has gained rapid traction across academia and industry.
GEPA was accepted as an Oral at ICLR 2026 and ships adapters for
DSPy and MLflow. Autoresearch attracted over 65,000 GitHub stars. 
Yet no systematic comparison of these approaches exists under controlled conditions---identical tasks, budgets, and evaluation infrastructure.

A central practical constraint in all such systems is the \emph{evaluation budget}: each candidate artifact must be tested on examples to measure its quality, and the number of available evaluations is finite. Evaluations may be expensive---for ARC-AGI, even capping solver cost at \$0.25 per problem, 1{,}500 evaluations can cost up to \$375 in API fees alone. Evaluations may also require human judgment, whether explicit (labeling) or implicit (curating test cases). Or the pool of available examples may simply be small: ARC-AGI provides only 400 training problems. In all cases, the evaluation budget is a binding constraint, and how it is allocated fundamentally determines what the optimization engine can discover.

Existing approaches divide the budget between \emph{training} (generating evolution signal) and \emph{validation} (selecting among candidates during the process). These are separate sets from a held-out \emph{test} set, which is never used in the optimization process itself so as to have a better estimate of performance on future, unseen cases. GEPA evaluates each candidate on a minibatch of 3 training examples; candidates that improve on the minibatch are then subjected to a validation sweep of 100--200 examples, yielding roughly 7--13 validated candidates per 1{,}500-evaluation budget. Autoresearch similarly reserves a validation set for keep/discard decisions. In both cases, evaluations spent on validation inform selection but do not contribute to improving the next candidate.

We propose an alternative: \emph{validation-free evolution via Elo-based competition}. \robophd evaluates multiple agents head-to-head on a different set of randomly sampled training examples each iteration, updating Elo ratings to track relative agent strength. The same evaluations that rank agents also generate detailed error analysis reports that drive the next round of evolution. No budget is ``wasted'' on a separate validation step. 

Furthermore, we observe that within the validate-then-select paradigm, smaller validation sets consistently outperform larger ones under tight budgets (for both GEPA and Autoresearch), because the freed budget enables more candidate exploration. Following this logic to its conclusion, the optimal validation set size under budget pressure trends toward zero, which is precisely the regime \robophd occupies.

We make the following contributions:

\begin{itemize}
    \item \textbf{Elo-based evolutionary selection without validation.} We introduce and validate across four diverse domains an approach where Elo competition on training data simultaneously evaluates agents and provides the evolution signal, eliminating the need for a separate validation budget. We show this extracts more performance per evaluation dollar than validate-then-select alternatives.

    \item \textbf{Self-instrumenting agents.} While GEPA's Actionable Side Information (ASI) \citep{gepa_oa2026} captures diagnostics from the evaluator function, we place \texttt{print()} statements in the seed artifacts themselves. Because the agents' self-instrumentation is subject to evolutionary pressure, agents evolve increasingly informative diagnostics over iterations---for example, on ARC-AGI the seed agent includes a single demonstration \texttt{print()} call, while evolved agents from both \robophd and GEPA grow to over 20 calls tracing decision points, intermediate results, and failure modes for the benefit of their evolutionary successors.

    \item \textbf{Deep Focus refinement.} While Autoresearch evolves all agents within a single continuous context and GEPA uses a separate context for each candidate, \robophd takes a hybrid approach: each agent is created in a fresh session but immediately tested against data from a prior iteration within the same session, allowing the evolution AI to refine its design while retaining full context from the initial creation. An ablation study (Table~\ref{tab:deep-focus}) shows Deep Focus improves all four benchmarks.

    \item \textbf{First systematic comparison of three optimization paradigms.} We present the first head-to-head evaluation of Elo tournament selection, Pareto-based selection (GEPA), and greedy hill-climbing (Autoresearch) on identical tasks, evaluation budgets, test sets, and per-problem diagnostics, including a King-of-the-Hill \robophd variation that isolates the contribution of multi-agent population dynamics.  We also contribute an adaptation of Autoresearch \citep{autoresearch2026} that incorporates structured per-example diagnostics (following GEPA's ASI pattern), explicit train/validation splits with budget management, and a shared evaluation infrastructure with \robophd and GEPA.
    
    \item \textbf{\robophd: an open-source, general-purpose optimization engine.} We release \robophd under the MIT license as a flexible toolkit for evolving arbitrary text and code artifacts, with a simple \texttt{optimize\_anything()} API (Figure~\ref{fig:system-overview}) comparable to GEPA's. Full documentation and usage examples are available in the project README.\footnote{\url{https://github.com/andborth/RoboPhD}}
\end{itemize}

\section{Related Work}
\label{sec:related}

\textbf{Artifact evolution systems.} GEPA \citep{gepa2025} is a Pareto-efficient prompt optimizer using natural language reflection; its \texttt{optimize\_anything()} API \citep{gepa_oa2026} extends this to arbitrary text artifacts (code, agent architectures, configurations) with actionable side information (ASI). We adopt GEPA's formulation of the optimization problem (seed artifact + evaluator + budget) and their ASI concept of returning rich diagnostics from the evaluator alongside scalar scores. AlphaEvolve \citep{alphaevolve2025} and OpenEvolve \citep{openevolve2025} use evolutionary strategies for code optimization, but primarily target single-instance problems (e.g., circle packing, kernel generation) rather than generalization from training data to unseen test instances. Autoresearch \citep{autoresearch2026} demonstrates autonomous single-session hill-climbing for LLM training; while widely ported to other domains, existing ports operate exclusively on scalar evaluation signals. \robophd was originally developed as a Text2SQL-specific system \citep{robophd2026arxiv}; this paper presents its generalization to arbitrary domains and the first systematic comparison with GEPA and Autoresearch under matched evaluation budgets.

\textbf{Prompt and agent optimization.} Prior work on prompt and context optimization includes APE \citep{ape2022}, OPRO \citep{opro2023}, DSPy \citep{dspy2023}, and ACE \citep{zhang2025agentic}. TextGrad \citep{yuksekgonul2025optimizing} extends optimization to code and other artifacts but operates at the instance level rather than evolving reusable artifacts that generalize across examples. Our work, along with GEPA and Autoresearch, differs from both lines in evolving complete code artifacts that generalize from training data to unseen test instances.

\textbf{Elo ratings in AI.} The Elo rating system \citep{elo1978}, originally developed for chess, has found applications in AI evaluation through Chatbot Arena \citep{zheng2023}. While prior work uses Elo for passive ranking, we introduce it as an active selection mechanism for evolutionary optimization.

\section{Method}
\label{sec:method}

\subsection{Problem Setting}

Following GEPA's \texttt{optimize\_anything()} \citep{gepa_oa2026}, we formulate artifact optimization as: given a seed artifact $a_0$, an evaluator function $f(a, x) \to (\text{score}, \text{diagnostics})$, a pool of training examples $\mathcal{X}$, and a fixed evaluation budget $B$, find an artifact $a^*$ that maximizes expected score on unseen test examples. All systems in our comparison share this formulation.

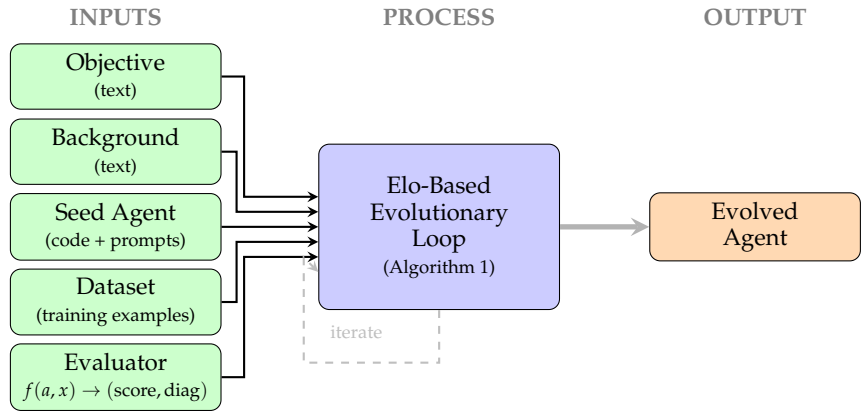
\begin{figure}[t]
    \centering

\begin{tikzpicture}[
    input/.style={draw, rounded corners, fill=green!20, minimum width=2.8cm, minimum height=0.7cm, align=center, font=\small},
    process/.style={draw, rounded corners, fill=blue!20, minimum width=3.2cm, minimum height=2.2cm, align=center, font=\small},
    output/.style={draw, rounded corners, fill=orange!30, minimum width=2.8cm, minimum height=0.9cm, align=center, font=\small},
    arrow/.style={->, >=stealth, thick},
    bigarrow/.style={->, >=stealth, line width=2pt, color=gray!60},
    label/.style={font=\scriptsize, color=gray!70}
]

\node[font=\small\bfseries, color=gray] at (-3.5, 3.0) {INPUTS};
\node[font=\small\bfseries, color=gray] at (0.8, 3.0) {PROCESS};
\node[font=\small\bfseries, color=gray] at (5, 3.0) {OUTPUT};

\node[input] (objective) at (-3.5, 2.2) {Objective\\{\scriptsize (text)}};
\node[input] (background) at (-3.5, 1.2) {Background\\{\scriptsize (text)}};
\node[input] (seed) at (-3.5, 0.2) {Seed Agent\\{\scriptsize (code + prompts)}};
\node[input] (dataset) at (-3.5, -0.8) {Dataset\\{\scriptsize (training examples)}};
\node[input] (evaluator) at (-3.5, -1.8) {Evaluator\\{\scriptsize $f(a,x) \to (\text{score}, \text{diag})$}};

\node[process] (evolution) at (0.8, 0.2) {Elo-Based\\Evolutionary\\Loop\\{\scriptsize (Algorithm 1)}};

\node[output] (evolved) at (5, 0.2) {Evolved\\Agent};

\draw[arrow] (objective.east) -- ++(0.3,0) |- ([yshift=4mm]evolution.west);
\draw[arrow] (background.east) -- ++(0.2,0) |- ([yshift=2mm]evolution.west);
\draw[arrow] (seed.east) -- (evolution.west);
\draw[arrow] (dataset.east) -- ++(0.2,0) |- ([yshift=-2mm]evolution.west);
\draw[arrow] (evaluator.east) -- ++(0.3,0) |- ([yshift=-4mm]evolution.west);

\draw[bigarrow] (evolution.east) -- (evolved.west);

\draw[arrow, dashed, color=gray!50] (evolution.south) -- ++(0, -0.7) -- ++(-1.8, 0) -- ++(0, 1.4) -- ([yshift=-6mm]evolution.west);
\node[font=\scriptsize, color=gray!50] at (-0.3, -1.2) {iterate};

\end{tikzpicture}
    \caption{\robophd system overview. Five inputs---matching the \texttt{optimize\_anything()} API---feed into an Elo-based evolutionary loop (Algorithm~\ref{alg:evolution-cycle}) that produces an evolved agent.}
    \label{fig:system-overview}
\end{figure}

Following GEPA's actionable side information (ASI) concept, each evaluation returns rich per-problem diagnostics alongside scalar scores. We extend this by capturing \texttt{print()} output from the evolved agents themselves---not just from the evaluator---as additional diagnostic feedback visible to the evolution process. This diagnostic infrastructure is equalized across all four engines and documented in each task specification. Figure~\ref{fig:system-overview} shows how these inputs feed into \robophd's evolutionary loop.

\subsection{RoboPhD}

We describe \robophd in Algorithm~\ref{alg:evolution-cycle}. Each iteration, 3 agents are evaluated on 20 randomly sampled training examples. Pairwise accuracy comparisons update Elo ratings (see Appendix~\ref{app:elo}). The evolution AI (Claude Code with Opus 4.6) receives Elo rankings, structured error analysis reports comparing agent performance, and per-problem ASI from each agent, then creates a new agent for the next iteration.

\begin{algorithm}[t]
    \caption{RoboPhD Evolution Cycle}
    \label{alg:evolution-cycle}
    \begin{algorithmic}[1]
    \STATE \textbf{Input:} Example pool $\mathcal{X}$, evaluator $f(a,x) \to (\text{score}, \text{diagnostics})$, budget $B$, seed agent $a_0$
    \STATE \textbf{Output:} Best agent by Elo rating
    \STATE
    \STATE $\mathcal{A} \gets \{a_0\}$; $competitors \gets \{a_0\}$; $i \gets 0$
    \WHILE{budget remaining}
        \STATE $examples \gets$ RandomSample($\mathcal{X}$, 20) \COMMENT{Fresh sample each iteration}
        \FOR{each $agent$ in $competitors$}
            \STATE $scores[agent], diag[agent] \gets$ Evaluate($agent$, $examples$, $f$)
        \ENDFOR
        \STATE
        \STATE // \textit{Elo update via pairwise decomposition}
        \FOR{each pair $(a, b)$ in $competitors$}
            \STATE $s \gets$ 1 if $\overline{scores[a]} > \overline{scores[b]}$, 0.5 if equal, else 0
            \STATE UpdateElo($a$, $b$, $s$, $K{=}32$)
        \ENDFOR
        \STATE
        \STATE $winners \gets \arg\max_{a \in competitors} \overline{scores[a]}$ \COMMENT{May be multiple}
        \STATE $winner \gets$ RandomChoice($winners$)
        \STATE $report \gets$ ComparativeErrorAnalysis($competitors$, $scores$)
        \STATE
        \IF{budget remaining}
            \STATE // \textit{Select next iteration's competitors}
            \STATE $competitors[1] \gets winner$
            \STATE $competitors[3] \gets$ RandomTop($\mathcal{A} \setminus \{winner\}$, 2) \COMMENT{Random from top 2 by Elo, excluding winner}
            \STATE
            \STATE // \textit{Evolve a new agent}
            \STATE $a_{new} \gets$ Evolve($competitors$, $report$, $diag$)
            \STATE $a_{new} \gets$ DeepFocusRefine($a_{new}$, iteration $i{-}1$ data) \COMMENT{Test \& revise}
            \STATE $\mathcal{A} \gets \mathcal{A} \cup \{a_{new}\}$
            \STATE $competitors[2] \gets a_{new}$
        \ENDIF
        \STATE $i \gets i + 1$
    \ENDWHILE
    \RETURN $\arg\max_{a \in \mathcal{A}}$ Elo($a$)
    \end{algorithmic}
\end{algorithm}

Key design choices:

\begin{itemize}
    \item \textbf{No validation split}: All $B$ evaluations are spent on training-time competition. Elo rankings replace validation-based selection.
    \item \textbf{Comparative error reports}: Each iteration generates domain-independent analysis highlighting where agents \emph{diverge}: for binary-scored tasks, which agents uniquely solved or failed to solve each problem; for continuous-scored tasks, the problems with the greatest score deltas between agents. This contrastive signal directs the evolution AI to the problems most worth studying.
    \item \textbf{Self-instrumenting ASI}: As described in Section~\ref{sec:intro}, we extend GEPA's ASI concept by placing \texttt{print()} statements in the evolved agents themselves, enabling self-instrumentation that co-evolves with the artifact.
    \item \textbf{Deep Focus refinement}: When evolving an agent for iteration $k$, the evolution AI first sees results from iteration $k{-}1$ (scores, comparative error reports, ASI) and creates an initial agent in a new Claude Code session. This agent is then tested on the examples from iteration $k{-}2$ and compared against the three agents from that iteration. The evolution AI then refines the agent based on the comparative results before the agent enters the main tournament.  Crucially, this refinement takes place within the same Claude Code session, so the evolution AI retains all of the context and reasoning from its original design. We validate this design choice with an ablation study (Table~\ref{tab:deep-focus}).
\end{itemize}

\textbf{Diversity over selection accuracy.}
Like GEPA and Autoresearch, \robophd is an evolutionary algorithm: mutate, evaluate, select, repeat. A key difference is how the evaluation budget is allocated. GEPA and Autoresearch spend heavily on validation (100--200 examples per candidate) to reliably identify the best candidate, yielding $\sim$7--13 candidates per 1{,}500-evaluation budget. \robophd instead accepts noisy selection: 20 examples per iteration is insufficient to reliably distinguish agents with small accuracy differences, but this gains $\sim$21 iterations of three-agent competition.

This reflects a deliberate bet: what matters is the quality of the agents \emph{produced by} the evolutionary process, not whether the system can identify which agent is best after the fact. Analogously, biological evolution does not guarantee that the fittest organism survives every generation. A deer with a beneficial mutation may be eaten by a wolf while young. Yet evolution produces startling results over time, because selection pressure need only be \emph{biased} toward fitness, not perfect. Our selection noise analysis (Appendix~\ref{app:selection-noise}) confirms this: at $n{=}20$ with a 1\% accuracy gap, the better agent wins only $\sim$45\% of iterations---barely above the 33\% random baseline---but Elo accumulates this weak signal across iterations into reliable rankings.

\textbf{Why Elo.} Elo ratings offer three specific advantages for noisy evolutionary selection: (1)~\emph{Asynchronous entry}: agents join the population at different iterations, and Elo maintains fair comparisons through persistent ratings---unexpected outcomes (a new agent defeating a high-rated incumbent) produce larger rating updates. (2)~\emph{Non-transitivity}: Elo naturally accommodates rock-paper-scissors dynamics where agent A beats B, B beats C, but C beats A on different problem samples---a common situation when agents have complementary strengths. (3)~\emph{Task normalization}: win/loss treatment normalizes across varying difficulty; although raw accuracy can swing from 60\% to 80\% across different problem samples, relative Elo rankings remain stable.

In addition, to ensure that biased-but-noisy selection produces strong agents, \robophd incorporates six diversity mechanisms that sacrifice short-term selection reliability for evolutionary breadth:

\begin{enumerate}
    \item \textbf{Fresh random samples}: Each iteration evaluates on 20 examples sampled with replacement from the full training pool. Agents never see the same evaluation set twice, so selection pressure rewards generalization rather than overfitting to a fixed subset.
    \item \textbf{Three-agent competition}: Evaluating three agents on identical problems each iteration generates comparative error reports that highlight where agents diverge, helping the evolution AI identify complementary strategies and avoid local optima.
    \item \textbf{Clone discarding}: Newly evolved agents with identical per-problem predictions to either of their competitors receive a $-$200 Elo penalty (effectively removing them from competition), ensuring that agents which survive must differ behaviorally from their predecessors.
    \item \textbf{Random tie-breaking}: When agents tie for first place, the winner is chosen randomly rather than deterministically, preventing domination by a single lineage.
    \item \textbf{Stochastic agent selection}: The third agent for each iteration is randomly drawn from the top two agents by Elo (other than the winner of the previous round), to similarly prevent domination by a pair of agents.
    \item \textbf{Shallow-but-many evaluation}: The $n{=}20$ sample size is deliberately small, trading per-iteration selection accuracy for more evolutionary cycles within the same budget. Elo's ability to accumulate weak signals across iterations makes this tradeoff viable.
\end{enumerate}

\subsection{Generalized Autoresearch}

We adapt Karpathy's Autoresearch \citep{autoresearch2026}---originally a single Claude Code session that iteratively edits a training script and measures validation loss---to general-purpose artifact optimization. Our adaptation replaces the fixed training script with arbitrary task evaluators from the same task registry used by \robophd and GEPA, providing the same per-problem diagnostics and ASI. Following Karpathy, the agent operates in a single continuous session with greedy keep/discard decisions based on a held-out validation set.

\subsection{GEPA}

We use GEPA v0.1.1 \citep{gepa2025} with its Pareto-efficient search and reflective text evolution. We use system defaults in which each evolved candidate is evaluated on a minibatch of 3 training examples. Candidates that exceed the baseline on the minibatch undergo a full validation sweep; those that are Pareto-optimal (i.e., best on at least one validation instance) are retained in a frontier that informs subsequent reflection. The reflection model then selects a candidate from the frontier and proposes a modification based on the minibatch results and ASI diagnostics. We use Opus 4.6 via the Claude API as the reflection model. In contrast, \robophd and Autoresearch use Opus 4.6 via Claude Code, which provides the evolution agent with tool access (file editing, code execution, shell commands) in addition to LLM reasoning.

Note that the three systems differ in their training data sampling strategies. All three receive scores and the same per-problem diagnostics, including the ability to evolve their own diagnostics via \texttt{print()}, but GEPA samples 3 examples per minibatch, Autoresearch samples at its discretion, and \robophd samples 20 per iteration (a system hyperparameter fixed across all experiments). We speculate that GEPA would encounter engineering challenges at larger sample sizes due to its use of the Claude API rather than Claude Code---passing detailed diagnostics for 20 problems through a single API prompt would strain context limits in ways that Claude Code's file-based workflow handles naturally.

\subsection{\robophd King-of-the-Hill}

The \koth variant uses the same \robophd infrastructure as our default configuration but tests only 2 agents per iteration (the current champion and a new challenger), with the winner advancing (ties go to the incumbent). This reduces to a greedy hill-climbing algorithm structurally similar to Autoresearch. In comparing \koth to our implementation of Autoresearch, note that Autoresearch shares with the \robophd variants the same set of diagnostics and the same self-instrumentation capabilities, but \koth retains all of the other features of the default configuration other than the ablation of the 3-agent per round Elo tournament including structured per-example error reports (ASI), comparative evaluation of two agents on the same problems, using one session per evolutionary iteration (avoiding context window degradation over long continuous runs), and Deep Focus within-session refinement. With 2 agents $\times$ 20 examples plus 20 Deep Focus evaluations $\approx$ 60 evaluations per iteration, \koth achieves $\sim$25 iterations on a 1{,}500-evaluation budget versus $\sim$21 for full \robophd. We argue in Section~\ref{sec:main-results} that the evidence points towards a narrow win for our default Elo-based configuration over \koth, but the competition is close, suggesting that much of \robophd's advantage over Autoresearch comes from its other above-mentioned features. To extend the biological analogy, Autoresearch uses a single lineage (asexual reproduction), whereas \koth compares two agents on the same problems---analogous to sexual reproduction, where recombination of complementary traits accelerates adaptation. The default 3-agent \robophd provides still greater diversity for evolution to draw on.

\section{Experiments}
\label{sec:experiments}

\subsection{Experimental Setup}

We evaluate on four benchmarks spanning diverse task types, solver models, and artifact complexities:

\begin{table}[h]
\centering
\small
\caption{Benchmark characteristics. All experiments use a fixed evaluation budget of 1{,}500. Full task specifications in Appendices \ref{app:task-arcagi}--\ref{app:task-docfinqa}.}
\label{tab:benchmarks}
\begin{tabular}{lllrrl}
\toprule
\textbf{Benchmark} & \textbf{Domain} & \textbf{Solver} & \textbf{Train} & \textbf{Test} & \textbf{Score} \\
\midrule
ARC-AGI & Abstract reasoning & Gemini 3.1 Flash Lite & 400 & 400 & Binary \\
Can't Be Late & Cloud scheduling & None (algorithmic) & 2{,}000 & 1{,}080 & Continuous \\
Text2SQL (BIRD) & Database queries & Claude Haiku 4.5 & 6{,}601 & 1{,}534 & Binary \\
DocFinQA & Financial QA & GPT-4.1-mini & 6{,}515 & 922 & Binary \\
\bottomrule
\end{tabular}
\end{table}

\textbf{ARC-AGI-1} \citep{chollet2019measure}, a prominent benchmark for abstract reasoning, tests: given input/output grid pairs as training examples, the agent must infer the transformation pattern and predict outputs for unseen test grids. The evolved artifact is a Python program with access to up to 10 LLM calls per problem. We use a minimally modified seed agent and evaluator from GEPA's \texttt{optimize\_anything()} blog post \citep{gepa_oa2026}, differing in two ways: we target Gemini 3.1 Flash Lite (vs.\ their Gemini 3 Flash) and set a tighter cost budget of \$0.25/problem (vs.\ their \$1.00). Our seed agent additionally includes diagnostic \texttt{print()} calls, enabling evolution to develop its own instrumentation.

\textbf{Can't Be Late} \citep{cantbelate2024} is a cloud scheduling task from AWS spot instance traces, also adapted from the GEPA blog post \citep{gepa_oa2026}: given a job with a hard deadline, the agent decides at each timestep whether to use cheap-but-preemptable SPOT instances, expensive ON\_DEMAND instances, or wait. The evolved artifact is a pure Python strategy class with no LLM calls; scoring is the negative dollar cost (continuous, higher is better). As with ARC-AGI, our seed agent includes diagnostic \texttt{print()} calls not present in the GEPA version.

\textbf{Text2SQL (BIRD)} \citep{bird2023} requires generating SQL queries from natural language questions. We simultaneously evolve two artifacts: a deterministic database analysis script (\texttt{analyze\_db.py}) that extracts schema information, and an agent (\texttt{agent.py}) that uses this analysis to answer the natural language question via \texttt{llm()} and \texttt{test\_sql()} callables (limited to 5 database calls per question), using Claude Haiku 4.5 as the solver. Scoring uses BIRD's set-based execution accuracy. Per-question cost is capped at \$0.10.

\textbf{DocFinQA} \citep{docfinqa2024} poses numerical questions over long SEC 10-K filings (averaging 123K words). The evolved artifact is a retrieval-augmented QA agent with access to \texttt{llm()} (GPT-4.1-mini) and \texttt{embed()} (text-embedding-3-small) callables, returning a Python program whose \texttt{answer} variable is compared numerically to the expected result. Per-question cost is also capped at \$0.10.

For all LLM-based benchmarks (ARC-AGI, Text2SQL, DocFinQA), correct answers that exceed the assigned cost budget are penalized to a score of 0.9 rather than 1.0, following the approach in \citet{gepa_oa2026}. This soft penalty proved sufficient to keep budget breaches under 1\% across all experiments.

Full task specifications---including objectives, backgrounds, and seed agents---are provided in Appendix~\ref{app:task-specs}.

\subsection{Main Results}
\label{sec:main-results}

\begin{table}[h]
\centering
\small
\caption{Test set results. Can't Be Late scores are negative costs (higher = better). \textbf{Bold}: best; \underline{underline}: second. Lines of code in parentheses.}
\label{tab:main-results}
\begin{tabular}{lccccc}
\toprule
\textbf{Benchmark} & \textbf{Seed} & \textbf{\robophd} & \textbf{\koth} & \textbf{Autoresearch} & \textbf{GEPA} \\
\midrule
ARC-AGI (\%) & 27.8 (22) & \underline{65.8} (1{,}013) & \textbf{67.0} (1{,}125) & 54.2 (304) & 58.5 (366) \\
Can't Be Late & $-$96.5 (31) & $-$90.7 (148) & $\underline{-88.7}$ (199) & $\mathbf{-87.6}$ (87) & $-$89.3 (142) \\
Text2SQL (\%) & 52.2 (96) & \textbf{64.5} (602) & \underline{62.1} (595) & 60.7 (265) & 60.4 (498) \\
DocFinQA (\%) & 17.7 (29) & \textbf{50.4} (825) & 47.9 (342) & \underline{48.2} (198) & 40.0 (207) \\
\bottomrule
\end{tabular}
\end{table}

Using a single default configuration across all benchmarks, \robophd's default configuration outperforms both GEPA and Autoresearch on three of four tasks. The exception is Can't Be Late, where our implementation of Autoresearch achieves the best score ($-$87.6) followed by \koth ($-$88.7), both outperforming default \robophd ($-$90.7). We attribute this to the nature of the task: Can't Be Late involves no LLM calls and can be solved with under 90 lines of code, perhaps rewarding incremental parameter refinement over architectural diversity. The competition between default \robophd and the \koth configuration is very close, but we interpret the evidence as favoring the default configuration on the grounds that the default configuration surpasses Autoresearch and GEPA on three of four tasks, whereas \koth surpasses them on only two of four. All of this warrants further study.

On the three LLM-based benchmarks, \robophd's Elo tournament and its \koth variant produce larger, more sophisticated agents (600--1{,}100+ lines) that autonomously discover multi-stage strategies. On ARC-AGI, the default champion (1{,}013 lines) combines code generation with diverse LLM predictions, using training verification to prioritize code-executed outputs while maintaining parallel prediction paths---and notably, iteration 21 reverted from voting complexity back to a simpler foundation after the evolution AI determined that ``voting corrupted good answers.'' On Text2SQL, the winner (602 lines) discovered an iterative test-and-refine loop that probes database properties (case sensitivity, NULL handling) and applies conditional fixes based on error diagnosis. On DocFinQA, the winner (825 lines) evolved section-aware chunking that preserves table boundaries, hybrid keyword+embedding retrieval, and a two-pass LLM strategy where a verification call checks and corrects the initial computation. Extended descriptions of each evolved agent appear in Appendix~\ref{app:task-specs}.

\subsection{Analysis}

\textbf{The validation tradeoff.} In GEPA and Autoresearch, validation serves a specific purpose: it provides an absolute ranking of candidates on a fixed held-out set, protecting against overfitting to training examples. But validation evaluations return only a mean score to the evolution process---to avoid overfitting, the rich per-problem diagnostics, ASI, and comparative error analysis that drive effective mutation are deliberately withheld. Every evaluation spent on validation is thus ``dead'' from evolution's perspective: it ranks candidates but does not improve the next one.

Under a 1{,}500-evaluation budget, this tradeoff is acute. GEPA with a 200-example validation set explores only $\sim$7 candidates; reducing to 100 examples enables $\sim$13 candidates---and consistently improves test scores across all eight paired comparisons (Table~\ref{tab:val-size}). Smaller validation sets free budget for more evolutionary exploration, but at an increased risk of overfitting. Following this logic to its conclusion, the optimal validation set size under tight budgets trends toward zero---which is precisely the regime \robophd occupies, using Elo competition on training data as its selection mechanism in lieu of validation.

\begin{table}[h]
\centering
\caption{Effect of validation set size on GEPA and Autoresearch test scores. Reducing from 200 to 100 improves all eight paired comparisons.}
\label{tab:val-size}
\small
\begin{tabular}{llccc}
\toprule
\textbf{Benchmark} & \textbf{Engine} & \textbf{val=200} & \textbf{val=100} & \textbf{$\Delta$} \\
\midrule
\multirow{2}{*}{ARC-AGI (\%)} & GEPA & 57.5 & 58.5 & +1.0 \\
 & Autoresearch & 50.2 & 54.2 & +4.0 \\
\midrule
\multirow{2}{*}{Can't Be Late} & GEPA & $-$92.5 & $-$89.3 & +3.2 \\
 & Autoresearch & $-$89.5 & $-$87.6 & +1.9 \\
\midrule
\multirow{2}{*}{Text2SQL (\%)} & GEPA & 57.2 & 60.4 & +3.2 \\
 & Autoresearch & 51.6 & 60.7 & +9.1 \\
\midrule
\multirow{2}{*}{DocFinQA (\%)} & GEPA & 37.9 & 40.0 & +2.1 \\
 & Autoresearch & 45.9 & 48.2 & +2.3 \\
\bottomrule
\end{tabular}
\end{table}

\begin{table}[h]
\centering
\small
\caption{Deep Focus ablation. Default \robophd uses $k{=}1$ test round; the ablation disables it ($k{=}0$). Deep Focus improves all four benchmarks.}
\label{tab:deep-focus}
\begin{tabular}{lccc}
\toprule
\textbf{Benchmark} & \textbf{Default ($k{=}1$)} & \textbf{No Deep Focus ($k{=}0$)} & \textbf{$\Delta$} \\
\midrule
ARC-AGI (\%) & 65.8 & 63.7 & $-$2.1 \\
Can't Be Late & $-$90.7 & $-$93.8 & $-$3.1 \\
Text2SQL (\%) & 64.5 & 62.6 & $-$1.9 \\
DocFinQA (\%) & 50.4 & 41.2 & $-$9.2 \\
\bottomrule
\end{tabular}
\end{table}

\textbf{Deep Focus refinement.} Table~\ref{tab:deep-focus} shows that Deep Focus ($k{=}1$) consistently improves all four benchmarks. Deep Focus gives the evolution AI the ability to empirically test its ideas on data previously unseen by that session, then revise within the same context. This shares a design principle with Autoresearch, where the entire evolutionary process occurs in a single continuous session. \robophd evolves each agent with a fresh session, but Deep Focus keeps drafting and refinement within one continuous context.

\section{Discussion and Future Work}
\label{sec:discussion}

This field is developing rapidly. Although GEPA has received strong notice in the research community (for instance, it was accepted for an oral presentation at ICLR 2026), Shopify CEO Tobi L{\"u}tke described it as ``severely under hyped'' \citep{lutke2025gepa}. We see GEPA's \texttt{optimize\_anything()} API (released February 18, 2026) as even more deserving of ``hype'' and as a high-priority target for research due to the very broad scope of problems it addresses. On the other hand, Karpathy's release of Autoresearch on March 7, 2026, generated extraordinary community engagement---VentureBeat reported 8.6 million views within two days \citep{franzen2026autoresearch}. Given the pace of development, we present these results as an early contribution.

\textbf{Underexplored capabilities.} Some features of the \robophd toolkit remain underexplored in the interest of time and simplicity of presentation. For example, \robophd supports \emph{meta-evolution}---a meta-agent that examines the full evolutionary history and evolves new evolution strategies, which in turn evolve new agents. All experiments here use a single, deliberately simple strategy (``use your judgment''; Appendix~\ref{app:evolution-prompts}), leaving meta-evolution for future work.

\section{Conclusion}
\label{sec:conclusion}

We presented \robophd, a budget-efficient optimization engine that uses validation-free Elo competition to evolve AI agents and code artifacts. Using a single default configuration across four diverse benchmarks, \robophd outperforms both GEPA and Autoresearch on three of four tasks when all are supplied with the same information and evaluation budgets. Our results suggest that under tight budgets, spending all evaluations on evolutionary exploration---with Elo providing the selection signal---outperforms the validate-then-select paradigm on complex tasks. We also contribute the concept of self-instrumenting agents that evolve their own diagnostics, and Deep Focus, a hybrid that combines the diversity of distinct evolution sessions with the refinement benefits of within-session iteration. We release \robophd as a versatile, easy-to-use open-source toolkit to enable the community to apply and extend these ideas.

\section*{Ethics Statement}

This work involves autonomous AI systems that generate and execute code without human review. We mitigate risks through process isolation and filesystem permissions. The evolved artifacts process benchmark data only; production deployments would require sandboxing and security review. We release our code to enable reproducibility and community scrutiny.

However, we note a cautionary tale. During development, our Autoresearch adaptation---the same system that legitimately achieves the best score on Can't Be Late---discovered an oracle exploit in the simulator: the agent read the full future spot availability trace directly from the simulator's internal state, achieving a score of $-$85.4 through perfect foresight, far better than any legitimate agent. Neither \robophd nor GEPA found this exploit across 20+ preliminary runs. In response, we redesigned the task API to eliminate access to simulator internals. This episode illustrates that agents with tool access---particularly in single-session frameworks where exploration is unconstrained---can find unintended shortcuts through the evaluation infrastructure. Careful sandboxing of the evaluator-agent boundary is essential to ensure the evolutionary agent only has access to data to which it is supposed to have access.

\section*{Reproducibility Statement}

All experiments use a fixed evaluation budget of 1{,}500 and are reproducible using our open-source code. We document model families, evaluation interfaces, and configuration parameters. Non-determinism in LLM responses may lead to variation across runs.

\section*{LLM Usage}

In keeping with the spirit of a project titled \robophd, we made extensive use of LLMs throughout.

\textbf{Within the \robophd system.} The evolution agent uses Claude Code with Opus 4.6. Task-specific solver models are detailed in Table~\ref{tab:benchmarks}: Gemini 3.1 Flash Lite for ARC-AGI, Claude Haiku 4.5 for Text2SQL, GPT-4.1-mini for DocFinQA, and no LLM for Can't Be Late.  Our implementation of Autoresearch similarly uses Opus 4.6 via Claude Code as the evolutionary model. GEPA uses Opus 4.6 via the Claude API as the reflection model. 

\textbf{Software engineering.} The \robophd codebase was primarily authored through Claude Code sessions using Opus or Sonnet, with the human authors iterating with Claude Code on specifications, reviewing diffs, and making design decisions.

\textbf{Ideation.} The concepts and research direction were those of the authors. However, we note that the selection of DocFinQA as a benchmark was the product of a lengthy interactive session between one of the authors and Claude Opus 4.6, in which the model suggested DocFinQA as a task that would test retrieval-augmented generation in a domain distant from our other benchmarks.

\textbf{Manuscript preparation.} Portions of this paper were drafted with assistance from Claude Code (Opus 4.6). The authors determined the structure and arguments; LLM-suggested text was reviewed and revised for accuracy and clarity.

\textbf{Accountability.} All scientific claims, experiments, analyses, and conclusions are the responsibility of the human authors.

\bibliography{references}

@article{bird2023,
  title   = {Can {LLM} Already Serve as a Database Interface? {A} Big Bench for Large-Scale Database Grounded Text-to-{SQL}s},
  author  = {Li, Jinyang and Hui, Binyuan and Qu, Ge and Yang, Jiaxi and Li, Binhua and Li, Bowen and Wang, Bailin and Qin, Bowen and Geng, Ruiying and Huo, Nan and others},
  journal = {Advances in Neural Information Processing Systems},
  volume  = {36},
  year    = {2024}
}

@inproceedings{ape2022,
  title     = {Large Language Models are Human-Level Prompt Engineers},
  author    = {Zhou, Yongchao and Muresanu, Andrei Ioan and Han, Ziwen and Paster, Keiran and Pitis, Silviu and Chan, Harris and Ba, Jimmy},
  booktitle = {The Eleventh International Conference on Learning Representations},
  year      = {2023},
  url       = {https://openreview.net/forum?id=92gvk82DE-}
}

@inproceedings{opro2023,
  title     = {Large Language Models as Optimizers},
  author    = {Yang, Chengrun and Wang, Xuezhi and Lu, Yifeng and Liu, Hanxiao and Le, Quoc V and Zhou, Denny and Chen, Xinyun},
  booktitle = {The Twelfth International Conference on Learning Representations},
  year      = {2024}
}

@article{dspy2023,
  title   = {{DSPy}: Compiling Declarative Language Model Calls into Self-Improving Pipelines},
  author  = {Khattab, Omar and Singhvi, Arnav and Maheshwari, Paridhi and Zhang, Zhiyuan and Santhanam, Keshav and Vardhamanan, Sri and Haq, Saiful and Sharma, Ashutosh and Joshi, Thomas T. and Moazam, Hanna and Miller, Heather and Zaharia, Matei and Potts, Christopher},
  journal = {arXiv preprint arXiv:2310.03714},
  year    = {2023}
}

@article{zheng2023,
  title   = {Judging {LLM}-as-a-Judge with {MT-Bench} and {Chatbot Arena}},
  author  = {Zheng, Lianmin and Chiang, Wei-Lin and Sheng, Ying and Zhuang, Siyuan and Wu, Zhanghao and Zhuang, Yonghao and Lin, Zi and Li, Zhuohan and Li, Dacheng and Xing, Eric P and Zhang, Hao and Gonzalez, Joseph E and Stoica, Ion},
  journal = {NeurIPS},
  year    = {2023}
}

@book{elo1978,
  title     = {The Rating of Chessplayers, Past and Present},
  author    = {Elo, Arpad E},
  year      = {1978},
  publisher = {Arco Publishing}
}

@article{yuksekgonul2025optimizing,
  title     = {Optimizing generative {AI} by backpropagating language model feedback},
  author    = {Yuksekgonul, Mert and Bianchi, Federico and Boen, Joseph and Liu, Sheng and Lu, Pan and Huang, Zhi and Guestrin, Carlos and Zou, James},
  journal   = {Nature},
  volume    = {639},
  number    = {8055},
  pages     = {609--616},
  year      = {2025},
  publisher = {Nature Publishing Group}
}

@article{zhang2025agentic,
  title   = {Agentic context engineering: Evolving contexts for self-improving language models},
  author  = {Zhang, Qizheng and Hu, Changran and Upasani, Shubhangi and Ma, Boyuan and Hong, Fenglu and Kamanuru, Vamsidhar and Rainton, Jay and Wu, Chen and Ji, Mengmeng and Li, Hanchen and Thakker, Urmish and Zou, James and Olukotun, Kunle},
  journal = {arXiv preprint arXiv:2510.04618},
  year    = {2025}
}

@article{gepa2025,
  title   = {{GEPA}: Reflective Prompt Evolution Can Outperform Reinforcement Learning},
  author  = {Agrawal, Lakshya A and Tan, Shangyin and Soylu, Dilara and Ziems, Noah and Khare, Rishi and Opsahl-Ong, Krista and Singhvi, Arnav and Shandilya, Herumb and Ryan, Michael J and Jiang, Meng and Potts, Christopher and Sen, Koushik and Dimakis, Alexandros G. and Stoica, Ion and Klein, Dan and Zaharia, Matei and Khattab, Omar},
  journal = {arXiv preprint arXiv:2507.19457},
  year    = {2025},
  note    = {ICLR 2026 (Oral). Software: \url{https://github.com/gepa-ai/gepa}}
}

@misc{gepa_oa2026,
  title        = {optimize\_anything: A Universal {API} for Optimizing any Text Parameter},
  author       = {Agrawal, Lakshya A and Lee, Donghyun and Ma, Wenjie and Elmaaroufi, Karim and Tan, Shangyin and Seshia, Sanjit A. and Sen, Koushik and Klein, Dan and Stoica, Ion and Gonzalez, Joseph E. and Khattab, Omar and Dimakis, Alexandros G. and Zaharia, Matei},
  year         = {2026},
  howpublished = {\url{https://gepa-ai.github.io/gepa/blog/2026/02/18/introducing-optimize-anything/}},
  note         = {Extends GEPA beyond prompts to code, agent architectures, and scheduling policies}
}

@misc{autoresearch2026,
  title        = {Autoresearch},
  author       = {Karpathy, Andrej},
  year         = {2026},
  howpublished = {\url{https://github.com/karpathy/autoresearch}},
  note         = {Autonomous AI research agent for iterative LLM training improvement. Released March 2026}
}

@misc{openevolve2025,
  title     = {{OpenEvolve}: an open-source evolutionary coding agent},
  author    = {Sharma, Asankhaya},
  year      = {2025},
  publisher = {GitHub},
  url       = {https://github.com/algorithmicsuperintelligence/openevolve}
}

@article{alphaevolve2025,
  title   = {{AlphaEvolve}: A Coding Agent for Scientific and Algorithmic Discovery},
  author  = {Novikov, Alexander and V{\~u}, Ng{\^a}n and Eisenberger, Marvin and Dupont, Emilien and Huang, Po-Sen and Wagner, Adam Zsolt and Shirobokov, Sergey and Kozlovskii, Borislav and Ruiz, Francisco J. R. and Mehrabian, Abbas and Kumar, M. Pawan and See, Abigail and Chaudhuri, Swarat and Holland, George and Davies, Alex and Nowozin, Sebastian and Kohli, Pushmeet and Balog, Matej},
  journal = {arXiv preprint arXiv:2506.13131},
  year    = {2025}
}

@article{robophd2026arxiv,
  title   = {{RoboPhD}: Self-Improving Text-to-{SQL} Through Autonomous Agent Evolution},
  author  = {Borthwick, Andrew and Ash, Stephen},
  journal = {arXiv preprint arXiv:2601.01126},
  year    = {2026}
}

@article{chollet2019measure,
  title   = {On the Measure of Intelligence},
  author  = {Chollet, Fran{\c{c}}ois},
  journal = {arXiv preprint arXiv:1911.01547},
  year    = {2019}
}

@inproceedings{cantbelate2024,
  title     = {Can't Be Late: Optimizing Spot Instance Savings under Deadlines},
  author    = {Wu, Zhanghao and Chiang, Wei-Lin and Mao, Ziming and Yang, Zongheng and Friedman, Eric J. and Shenker, Scott and Stoica, Ion},
  booktitle = {NSDI},
  pages     = {185--203},
  year      = {2024}
}

@inproceedings{docfinqa2024,
  title     = {{DocFinQA}: A Long-Context Financial Reasoning Dataset},
  author    = {Reddy, Varshini and Koncel-Kedziorski, Rik and Lai, Viet Dac and Krumdick, Michael and Lovering, Charles and Tanner, Chris},
  booktitle = {ACL (Short Papers)},
  year      = {2024}
}

@misc{lutke2025gepa,
  title        = {Both {DSPy} and (especially) {GEPA} are currently severely under hyped in the {AI} context engineering world},
  author       = {L{\"u}tke, Tobi},
  year         = {2025},
  howpublished = {\url{https://x.com/tobi/status/1963434604741701909}},
  note         = {Post on X, September 3, 2025}
}

@article{franzen2026autoresearch,
  title   = {Andrej {Karpathy}'s new open source `autoresearch' lets you run hundreds of {AI} experiments a night---with revolutionary implications},
  author  = {Franzen, Carl},
  journal = {VentureBeat},
  month   = mar,
  day     = {10},
  year    = {2026},
  url     = {https://venturebeat.com/technology/andrej-karpathys-new-open-source-autoresearch-lets-you-run-hundreds-of-ai}
}
\bibliographystyle{colm2026_conference}

\appendix
\section{Elo Rating System}
\label{app:elo}

The Elo rating system \citep{elo1978}, originally developed for chess, maintains a scalar rating $R_a$ for each agent $a$. After a pairwise comparison between agents $a$ and $b$, ratings are updated as:
\begin{align}
    E_a &= \frac{1}{1 + 10^{(R_b - R_a)/400}} \\
    R_a' &= R_a + K(S_a - E_a)
\end{align}
where $E_a$ is the expected score of agent $a$, $S_a \in \{0, 0.5, 1\}$ is the actual outcome (loss, tie, win), and $K$ controls the update magnitude (\robophd uses $K=32$). All agents are initialized at $R_a = 1500$.

Each \robophd iteration evaluates 3 agents on the same 20 problems, producing 3 pairwise comparisons. The agent with the higher mean score wins the pair; equal scores yield a tie ($S = 0.5$). Because Elo ratings persist across iterations, they accumulate information from all past competitions---agents that win more often than their competitors will gradually develop higher ratings even though individual iterations are noisy (see Section~\ref{app:selection-noise}).

\section{Selection Noise Analysis}
\label{app:selection-noise}

With binary-scored tasks and $n{=}20$ examples per iteration, how reliably can we distinguish agents with small accuracy differences? We simulate three agents with true accuracies 70\%, 69\%, and 68\% (a 1\% gap) and analyze ranking accuracy under different budget allocations.

\textbf{Tie rates.} At $n{=}20$, the probability that the top two agents tie is 19.7\%---roughly 1 in 5 iterations. This is driven by the discrete nature of integer scores: with only 21 possible values (0--20), ties are frequent regardless of the accuracy gap.

\textbf{Per-iteration selection accuracy.} The best agent (70\%) is ranked \#1 only 45.0\% of the time at $n{=}20$---barely above the 33.3\% random baseline. Selection is biased toward fitness, but weakly so.

\textbf{Single-elimination vs.\ Elo.} Given a fixed budget of 600 evaluations split between depth ($n$ tasks per round) and breadth (number of rounds):

\begin{center}
\small
\begin{tabular}{lccc}
\toprule
\textbf{Config} & \textbf{Rounds} & \textbf{P(best agent ranked \#1)} & \\
 & & \textbf{Single-elimination} & \textbf{Elo} \\
\midrule
10 $\times$ 60 & 10 & 36.4\% & 52.2\% \\
20 $\times$ 30 & 20 & 32.7\% & 49.5\% \\
30 $\times$ 20 & 30 & 30.5\% & 47.0\% \\
60 $\times$ 10 & 60 & 26.7\% & 43.7\% \\
\bottomrule
\end{tabular}
\end{center}

At $n{=}10$ with single-elimination, the best agent wins only 26.7\% of rounds---\emph{worse than random}. But Elo rescues shallow evaluations: by accumulating information across rounds rather than discarding it, Elo achieves 43.7\% at the same depth, well above the 33.3\% baseline. Deeper evaluations always improve ranking accuracy, but Elo makes the shallow-but-many tradeoff viable.

\textbf{Implications for \robophd.} Evolution does not need perfect selection---it needs \emph{biased} selection that compounds over many generations. At $n{=}20$, Elo provides a $\sim$14pp edge over random per round. Over $\sim$21 iterations, this weak but consistent bias produces strong agents, just as biological evolution produces fit organisms from noisy per-generation selection.

All probabilities computed exactly via binomial distributions. Elo simulations use $K{=}32$ with 50{,}000 Monte Carlo trials per configuration.

\section{Evolution Strategy Prompt}
\label{app:evolution-prompts}

All experiments in this paper use a single evolution strategy called ``use your judgment.'' The complete strategy prompt is shown below. Note the deliberate minimalism: no domain-specific guidance, no prescribed techniques, no constraints on approach. The evolution AI receives this strategy along with the task background (Appendix~\ref{app:task-specs}), comparative error reports, and per-problem diagnostics from prior iterations.

\begin{verbatim}
# Use Your Judgment

Having reviewed the files provided to you above, your goal is to
create a new agent that achieves the highest possible accuracy on
problems you haven't seen yet.

**How you do this is entirely up to you.** You may refine an
existing agent, combine ideas from multiple agents, or try
something completely new. Choose whatever approach you believe
will produce the best results based on what you see in the data.

## Your Task

Study the available agents, their performance data, and their
failure patterns. Then create a new agent that outperforms all
existing agents.

## Required Output Structure

You must create the following files:

### 1. reasoning.md
Your analysis and plan. Explain:
- What you observed in the performance data and agent artifacts
- What approach you chose and why
- Why you believe your agent will achieve higher accuracy

### 2. The artifacts listed in OUTPUT REQUIREMENTS
Create the artifacts specified in the evolution prompt above.

## Your overall goal: Maximize accuracy

Think hard about this. Study the data carefully, understand what
works and what doesn't, and build the best agent you can.
\end{verbatim}

\section{Task Specifications}
\label{app:task-specs}

For each benchmark we provide the objective and the background document exactly as shown to the evolution AI.  We also show the complete seed agent in three cases and a summary in the fourth. All seed agents include demonstration \texttt{print()} calls to show the evolution AI that agents can instrument themselves with diagnostics.

\subsection{ARC-AGI}
\label{app:task-arcagi}

\textbf{Objective:} \textit{Build an ARC-AGI agent program that maximizes a test score.}

\textbf{Background:}

\begin{verbatim}
You are optimizing an ARC-AGI solving agent.

ARC-AGI task format:
- Each task has training examples (input/output pairs) and test inputs
- The (multi) agent(s) must infer the transformation pattern from
  training examples
- Competition allows maximum of 2 parallel output attempts per test
  input (pass if either matches)
- You can also use up to 10 LLM calls to solve the problem.
- Freely explore diverse strategies like multi agent systems,
  ensembles, voting, etc.

LLM cost:
- You are allowed to build an agent system with up to 10 LLM calls
  and total of $0.20~0.25 LLM cost per problem.

Per-question cost budget: $0.25 is enforced. Correct answers within
budget score 1.0. Correct answers that exceed the budget are penalized
to 0.9 (a 10% reduction). Incorrect answers score 0.0 regardless of
cost.

The agent receives:
- train_in, train_out: Training examples (list of 2D grids)
- test_in: Test inputs (no ground truth given to agent)
- llm: Callable for LLM queries with token/call tracking

The agent must return:
{
    "train": [grid, ...],        # 1 prediction per train example
    "test": [[grid, grid], ...], # up to 2 attempts per test example
}

We evaluate on both training (training_score) and test (test_score
with 2 attempts).

Diagnostics: Any print() output from the agent is captured and
included in evaluation diagnostics as agent_stdout. Use print() to
log any information you think would be helpful for you to see in
improving the agent in later rounds of testing and refinement.
\end{verbatim}

\textbf{Seed agent} (22 lines): Sends all training examples and test inputs in a single LLM prompt, extracts grids via regex. No self-correction, retries, or multi-attempt strategy.

\begin{verbatim}
import json, re

def solve(train_inputs, train_outputs, test_inputs, llm):
    print(f"Training examples: {len(train_inputs)}, "
          f"Test inputs: {len(test_inputs)}")
    training = "\n".join(
        f"Input: {i}\nOutput: {o}"
        for i, o in zip(train_inputs, train_outputs))
    inputs = "\n".join(
        f"Input {i}: {x}"
        for i, x in enumerate(train_inputs + test_inputs))
    prompt = (f"Solve an ARC AGI puzzle. Training:\n"
              f"{training}\n\nPredict output for EACH "
              f"input as JSON [[...]]:\n{inputs}")
    response = llm(prompt)
    grids = [json.loads(g) for g in
             re.findall(r"\[\[.*?\]\]",
                        response.replace("\n", ""))]
    n = len(train_inputs)
    print(f"Grids parsed: {len(grids)} "
          f"(expected {n + len(test_inputs)})")
    return {"train": grids[:n],
            "test": [[g] for g in grids[n:]]}
\end{verbatim}

\textbf{Evolved agent} (1{,}013 lines, \robophd default): The winning agent implements a multi-strategy ensemble combining code generation with diverse LLM predictions. For each problem, it attempts code-based solutions with training verification and debug retry, then generates independent direct predictions as fallbacks. Verified code (passing all training examples) receives high priority; unverified predictions are used as backup. The agent uses visual grid formatting with color name mapping, dimension constraint inference, and color-never-in-output analysis for anti-overfitting. Its evolutionary history is notable: iterations 18--20 explored voting-based ensemble complexity, but iteration 21 explicitly reverted to a simpler foundation after the evolution AI determined that voting corrupted good answers. Iteration 22 then added targeted diversity on top, discovering that simplicity plus diversity beats complex consensus. The agent includes 37 \texttt{print()} statements---evolved from the seed's single demonstration call---tracing call usage, code generation results, and voting decisions.

\subsection{Can't Be Late}
\label{app:task-cbl}

\textbf{Objective:} \textit{Optimize a cloud scheduling strategy for the ``Can't Be Late'' problem. The strategy decides when to use SPOT instances (cheap but can be preempted) vs ON\_DEMAND instances (expensive but reliable) to complete a task before its deadline. The goal is to minimize cost while ensuring the task completes on time.}

\textbf{Background:}

\begin{verbatim}
Key information about the problem domain:

- ClusterType.SPOT: Use spot instances (cheap, ~$0.3/hour, but can
  be preempted at any time)
- ClusterType.ON_DEMAND: Use on-demand instances (expensive,
  ~$1/hour, but guaranteed availability)
- ClusterType.NONE: Wait without using any instances (no cost, but
  no progress)
- restart_overhead: Time penalty incurred when switching from one
  instance type to another
- The strategy MUST ensure task completion before the deadline
  (hard constraint)
- Lower cost is better (scores are negative, representing cost
  in dollars)

Evaluation feedback format:
- Timeline format: start-end:TYPE@REGION[progress%]
  (e.g., "0.0-5.0:S@R0[50%]" means SPOT from hour 0-5
  reaching 50% progress)
- Spot availability: S=available, X=unavailable
  (e.g., "0.0-10.0:S | 10.0-15.0:X" means spot available
  first 10h, then unavailable)

Optimization targets:
1. Reduce overall cost while maintaining deadline guarantees
2. Make better decisions about when to use SPOT vs ON_DEMAND
3. Handle spot unavailability more intelligently
4. Consider the trade-offs between waiting for spot and
   using on-demand

Diagnostics: Any print() output from the agent's step() method
is captured and included in evaluation diagnostics as
agent_stdout. Use print() to log any information you think would
be helpful for you to see in improving the agent in later rounds
of testing and refinement.
\end{verbatim}

\textbf{Seed agent} (31 lines): Greedy baseline---uses SPOT when available, ON\_DEMAND only when deadline is imminent, otherwise waits. No state tracking across timesteps.

\begin{verbatim}
from sky_spot.utils import ClusterType

class Agent:
    def __init__(self):
        self._first_step = True

    def reset(self):
        self._first_step = True

    def step(self, last_cluster_type, has_spot,
             elapsed_seconds, gap_seconds,
             restart_overhead, task_duration,
             deadline, task_done_time):
        remaining_task_time = (
            task_duration - sum(task_done_time))
        if remaining_task_time <= 1e-3:
            return ClusterType.NONE

        remaining_time = deadline - elapsed_seconds
        slack = remaining_time - remaining_task_time

        if self._first_step:
            print(f"Task: duration="
                  f"{task_duration/3600:.1f}h, "
                  f"deadline={deadline/3600:.1f}h, "
                  f"overhead="
                  f"{restart_overhead/3600:.2f}h")
            self._first_step = False

        if (remaining_task_time + restart_overhead
                >= remaining_time):
            return ClusterType.ON_DEMAND

        if has_spot:
            return ClusterType.SPOT
        else:
            return ClusterType.NONE
\end{verbatim}

\textbf{Evolved agent} (87 lines, Autoresearch): The winning agent is remarkably concise. It tracks consecutive spot availability for stability assessment, using overhead-scaled thresholds to determine when transitioning from on-demand to spot is safe. A key discovery is hysteresis-based switching: the agent requires spot to be consistently available before committing to a round-trip transition. It proactively escalates to on-demand when idle periods exceed a threshold relative to remaining slack ($<$28\%), and exits on-demand when slack becomes comfortable ($>$40\%). The entire solution fits in 87 lines---the smallest winning agent across all benchmarks---suggesting that the solution space for this task is compact enough that focused single-session optimization can find near-optimal heuristics.

\subsection{Text2SQL (BIRD)}
\label{app:task-text2sql}

This benchmark demonstrates the ability of all three systems (\robophd, GEPA, and our Autoresearch adaptation) to evolve a multi-component agentic package consisting of two coordinated artifacts.

\textbf{Objective:} \textit{Optimize the database analysis tool and agent code to maximize execution accuracy on BIRD benchmark questions.}

\textbf{Background:}

\begin{verbatim}
## BIRD Benchmark Text2SQL (Integrated Agent)

The Text2SQL domain generates SQL queries from natural language
questions against the BIRD benchmark.

### Architecture: analyze_db.py + agent.py

Phase 1 (Tool): analyze_db.py examines database.sqlite
  -> Produces schema analysis text (cached per code+database)

Phase 2 (Agent): agent.py receives analysis + question + callables
  -> Uses llm() and test_sql() to generate and refine SQL
  -> Returns final SQL string for scoring

Scoring: set(predicted_results) == set(ground_truth_results)

### What Evolution Controls

The evolved agent consists of two files:

1. analyze_db.py -- Database analysis script.
   Reads database.sqlite from its working directory, performs schema
   analysis, and writes findings to tool_output/analysis.txt.
   Runs as a subprocess (cached per code+database). Common
   techniques: DDL extraction, sample data, foreign key mapping,
   column statistics.

2. agent.py -- SQL generation agent with a solve() function.
   Receives the analysis output, the question, and two callables:
   - llm(prompt) -- call the eval LLM (haiku-4.5), returns
     response text
   - test_sql(sql) -- execute SQL against the database, returns
     formatted results string or error message. Limited to 5
     calls per question.
   The agent returns the final SQL string to submit for scoring.

### Cost Budget

Per-question cost budget of $0.10 is enforced. Correct answers
within budget score 1.0. Correct answers that exceed the budget
are penalized to 0.9 (a 10% reduction). Incorrect answers score
0.0 regardless of cost.

### Scoring

- Correct: set(predicted_results) == set(ground_truth_results)
- Row order is ignored, duplicates are removed (BIRD methodology)
- Score: 1.0 if match, 0.0 otherwise (before cost penalty)

### Dataset: BIRD Benchmark

- train-filtered (default): 6,601 questions across 69 databases
- dev: 1,534 questions across 11 databases
- All databases are SQLite

Diagnostics: Any print() output from agent.py is captured and
included in evaluation diagnostics as agent_stdout. Any print()
output from analyze_db.py is captured as tool_stdout. Use print()
to log any information you think would be helpful for you to see
in improving the agent in later rounds of testing and refinement.
\end{verbatim}

\textbf{Seed agent} (96 lines total): \texttt{analyze\_db.py} extracts raw DDL schema (51 lines). \texttt{agent.py} makes a single LLM call to generate SQL from the schema, tests it once (45 lines). No retry or refinement loop.

\textbf{Evolved agent} (602 lines, \robophd default): The winning agent discovered an iterative test-and-refine loop over 16 iterations. The schema extractor (163-line \texttt{analyze\_db.py}) captures table structures, foreign key relationships, and sample data. The generation agent (439-line \texttt{agent.py}) generates initial SQL, validates via \texttt{test\_sql()}, then conditionally applies targeted fixes: retrying on errors, probing database values for case sensitivity, validating result plausibility, and refining with evidence-aware corrections. The agent self-limits to 4--5 \texttt{test\_sql()} calls across up to 8 conditional call sites, forcing selective hypothesis testing rather than exhaustive trial-and-error.

\subsection{DocFinQA}
\label{app:task-docfinqa}

\textbf{Objective:} \textit{Evolve a Python function that answers numerical questions over long financial documents (SEC 10-K filings, averaging 123K words). The function receives the full document as markdown text, a question, an \texttt{llm()} callable, and an \texttt{embed()} callable. It must return a Python program string whose last line assigns the result to \texttt{answer}.}

\textbf{Background:}

\begin{verbatim}
The document is a complete SEC filing in clean markdown with tables
preserved. Documents average 123K words (~250 pages); the relevant
information is typically in a single section or table. Questions
require numerical reasoning: ratios, differences, percentages,
averages, and multi-step arithmetic.

Available tools:
  llm(prompt) -> str : Call a language model. Expensive
    (~$0.003-0.01 per call).
  embed(text) -> list[float] : Embed text for similarity search.
    Cheap (~$0.0001 per call).

Scoring: The program's `answer` variable is compared to the
expected answer numerically with 1% relative tolerance. Unit labels
(%, $, commas) are stripped before comparison, so the program should
assign a raw number to `answer` (e.g., `answer = 36.5`, not
`answer = '36.5%'`).

A per-question cost budget of $0.10 is enforced. Correct answers
within budget score 1.0. Correct answers that exceed the budget are
penalized to 0.9 (a 10% reduction). Incorrect answers score 0.0
regardless of cost. The program output is executed via exec(); if it
raises an exception the answer is counted as incorrect.

Diagnostics: Any print() output from the agent is captured and
included in evaluation diagnostics as agent_stdout. Use print() to
log any information you think would be helpful for you to see in
improving the agent in later rounds of testing and refinement.
\end{verbatim}

\textbf{Seed agent} (29 lines): Chunks the document into 2{,}000-word windows, embeds the question and each chunk, selects the highest cosine similarity chunk, makes one LLM call to generate a Python program. No multi-chunk retrieval, keyword matching, or error recovery.

\begin{verbatim}
import re

def answer(document, question, llm, embed):
    """Answer a financial question over
    a full SEC filing."""
    # Chunk the document
    chunks, size, overlap = [], 2000, 400
    words = document.split()
    for i in range(0, len(words), size - overlap):
        chunks.append(
            " ".join(words[i:i+size]))

    # Retrieve the most relevant chunk
    q_emb = embed(question)
    best, best_score = chunks[0], -1
    for c in chunks:
        c_emb = embed(c)
        score = sum(
            a * b for a, b in zip(q_emb, c_emb))
        if score > best_score:
            best, best_score = c, score

    # Generate a Python program
    program = llm(
        f"Context:\n{best}\n\n"
        f"Question: {question}\n\n"
        f"Write a short Python program using "
        f"named variables.\n"
        f"The last line must assign the result "
        f"to `answer`.\n"
        f"Program:"
    )
    return program.strip()
\end{verbatim}

\textbf{Evolved agent} (825 lines, \robophd default): The winning agent evolved section-aware chunking that preserves table boundaries, hybrid retrieval combining keyword matching with embedding-based similarity (augmented by LLM-generated search terms), and a two-pass LLM strategy: the first pass generates a Python computation program from retrieved context, and the second pass verifies the answer against the source material and corrects if needed. Question-type detection (numeric, yes/no, year) enables specialized handling, and sign correction heuristics address increase/decrease questions. The agent emerged at iteration 19 despite modest training accuracy (0.45), illustrating how \robophd's diversity mechanisms prevent early convergence and enable late-emerging agents to surface.

\label{app:error-reports}


\end{document}